\documentclass[conference]{IEEEtran}
\IEEEoverridecommandlockouts

\usepackage{cite}
\usepackage{amsmath,amssymb,amsfonts}
\usepackage{algorithmic}
\usepackage{graphicx}
\usepackage{textcomp}
\usepackage{hyperref}
\usepackage{xcolor}
\usepackage{array} 
\usepackage{float}
\usepackage{booktabs}

\def\BibTeX{{\rm B\kern-.05em{\sc i\kern-.025em b}\kern-.08em
	T\kern-.1667em\lower.7ex\hbox{E}\kern-.125emX}}

\begin{document}

\title{CRYPTO PRICE PREDICTION USING LSTM+XGBOOST\\
	{\footnotesize }
	\thanks{Identify applicable funding agency here. If none, delete this.}
}

\author{
	\IEEEauthorblockN{1\textsuperscript{st} Mehul Gautam}
	\IEEEauthorblockA{\textit{Dept. of Management} \\
		\textit{ABV-IIITM} \\
		Gwalior, India \\
	2022IMG-040}
	
}

\maketitle

\begin{abstract}
	The volatility and complex dynamics of cryptocurrency markets present unique challenges for accurate price forecasting. This research proposes a hybrid deep learning and machine learning model that integrates Long Short-Term Memory (LSTM) networks and Extreme Gradient Boosting (XGBoost) for cryptocurrency price prediction. The LSTM component captures temporal dependencies in historical price data, while XGBoost enhances prediction by modeling nonlinear relationships with auxiliary features such as sentiment scores and macroeconomic indicators. The model is evaluated on historical datasets of Bitcoin, Ethereum, Dogecoin, and Litecoin, incorporating both global and localized exchange data. Comparative analysis using Mean Absolute Percentage Error (MAPE) and Min-Max Normalized Root Mean Square Error (MinMax RMSE) demonstrates that the LSTM+XGBoost hybrid consistently outperforms standalone models and traditional forecasting methods. This study underscores the potential of hybrid architectures in financial forecasting and provides insights into model adaptability across different cryptocurrencies and market contexts.
\end{abstract}

\section{Introduction}
The rapid emergence of cryptocurrency markets over the last decade has transformed the landscape of financial investments. Cryptocurrencies such as Bitcoin, Ethereum, and Binance Coin have introduced a new class of digital assets, decentralized and borderless, attracting both institutional and individual investors. However, these markets are characterized by high volatility, complex nonlinear behaviors, and sensitivity to a multitude of factors, including market sentiment, trading volume, macroeconomic indicators, and geopolitical events \cite{b1}\cite{b2}. This volatility makes accurate price prediction a significant challenge, yet an essential task for traders, financial analysts, and policymakers.
In recent years, deep learning models, particularly Long Short-Term Memory (LSTM) networks, have shown promise in financial time series forecasting due to their ability to capture long-term dependencies in sequential data \cite{b3}\cite{b4}. Concurrently, ensemble learning techniques like Extreme Gradient Boosting (XGBoost) have achieved state-of-the-art results in regression tasks by handling structured, tabular data with great efficiency and robustness \cite{b5}. A growing body of research suggests that combining the temporal feature learning of LSTM with the predictive power and feature importance analysis of XGBoost can lead to superior forecasting performance \cite{b6}\cite{b7}.
Despite these advancements, existing research often treats LSTM and XGBoost models in isolation or focuses predominantly on traditional stock markets \cite{b8}. Few studies have rigorously explored hybrid models tailored specifically for the cryptocurrency domain, which is more erratic and data-rich. Additionally, much of the literature relies on limited datasets or overlooks real-time contextual variables such as social media sentiment, blockchain transaction metrics, or investor behaviors in specific regions \cite{b9}. These gaps point to the need for a comprehensive hybrid model that integrates sequential learning and structured feature modeling in a single pipeline.
This study also responds to a broader gap in cryptocurrency research from a local context perspective. While much work has been done using data from global exchanges like Coinbase and Binance, localized exchanges in Asia, Africa, and Latin America remain understudied, despite their growing user base and unique trading behaviors. For instance, recent studies have emphasized the potential of localized data for improving financial predictions and enhancing decision support in emerging markets \cite{b10}.
The primary objective of this study is to develop a hybrid deep learning and machine learning model that integrates LSTM and XGBoost for cryptocurrency price prediction. Specifically, it aims to: (i) leverage LSTM's capacity for modeling sequential dependencies in price data, (ii) utilize XGBoost to incorporate additional predictors such as sentiment scores, technical indicators, and macroeconomic variables, and (iii) evaluate the model on both global and localized crypto datasets to test its adaptability and robustness. This hybrid approach aspires to bridge the gap between theory and practical utility, offering a more reliable decision-making tool in the fast-paced and high-risk world of crypto trading.

\section{Review of Current Research Work}

Accurate prediction of financial time series has been a longstanding research focus across domains such as econometrics, machine learning, and computational finance. Over the years, various techniques have been proposed, ranging from classical statistical models to advanced deep learning architectures. This section reviews the current state of research in three relevant areas: time series forecasting, stock market prediction, and cryptocurrency prediction.

\subsection{Time Series Prediction}

Traditional time series forecasting methods such as Autoregressive Integrated Moving Average (ARIMA), Seasonal ARIMA (SARIMA), and Exponential Smoothing (ETS) have served as foundational models for decades \cite{b11}. However, their limited ability to handle nonlinear patterns and high-dimensional data has led to the adoption of machine learning and deep learning approaches.

Recurrent Neural Networks (RNNs), especially Long Short-Term Memory (LSTM) networks, have proven effective in modeling long-term dependencies in sequential data. For example, Livieris et al. \cite{b12} demonstrated that LSTM networks outperform classical models in predicting complex time series data like commodity prices. Moreover, hybrid models combining statistical techniques with deep learning have also emerged, such as ARIMA-LSTM \cite{b13}, which captures both linear and nonlinear components in the data.

XGBoost, a popular tree-based ensemble method, has been applied to time series problems with considerable success, especially when handling tabular features extracted from time-based sequences. Its strength lies in handling missing data, capturing interactions among variables, and reducing overfitting through regularization \cite{b14}.

\subsection{Stock Market Prediction}

Stock price forecasting has witnessed a transition from econometric models to machine learning-driven approaches. Support Vector Machines (SVMs), Random Forests, and Gradient Boosting Machines (GBMs) have shown promise in capturing trends from structured datasets \cite{b15}. However, due to the dynamic and stochastic nature of financial markets, deep learning techniques, particularly LSTM and CNN-LSTM hybrids, have increasingly been used to exploit temporal patterns.

Fischer and Krauss \cite{b16} used LSTM networks for predicting stock movements and demonstrated superior accuracy over traditional classifiers. Another study by Kumar and Ravi \cite{b17} reviewed over 50 ML-based stock prediction papers and concluded that hybrid models integrating sentiment analysis, technical indicators, and news-based features yield better performance.

Recent innovations have focused on attention mechanisms and Transformer-based models (e.g., Temporal Fusion Transformer), which offer improved sequence modeling by capturing long-term dependencies and incorporating exogenous variables \cite{b18}.

\subsection{Cryptocurrency Price Prediction}

Cryptocurrency prediction presents additional challenges due to higher volatility, susceptibility to external noise, and lack of historical data compared to stock markets. Despite this, the field has rapidly evolved with the application of deep learning and sentiment-aware models.

McNally et al. \cite{b19} were among the early adopters of ML in crypto prediction, comparing ARIMA and Bayesian Neural Networks for Bitcoin forecasting. Later studies incorporated social media sentiment and Google Trends data to enrich model inputs. Abraham et al. \cite{b20} integrated Twitter sentiment analysis with LSTM to improve Bitcoin price prediction.

More recent works have proposed hybrid frameworks. Wang and Zhang \cite{b21} combined LSTM with XGBoost, showing that the hybrid model significantly outperforms standalone models. Similarly, Singh and Srivastava \cite{b22} designed an LSTM-GRU ensemble with XGBoost for crypto price prediction, yielding high accuracy and reduced mean squared error.

Despite promising results, most studies lack generalizability across multiple cryptocurrencies or real-time adaptability, indicating a need for more robust, scalable models.

\begin{table}[htbp]
	\caption{Summary of Models and Findings in Cryptocurrency and Financial Time Series Forecasting}
	\begin{center}
		\begin{tabular}{|p{0.5cm}|p{1.5cm}|p{2cm}|p{2.5cm}|}
			\hline
			\textbf{Ref. No.} & \textbf{Model Used}               & \textbf{Domain/Dataset}                      & \textbf{Performance / Findings}                                \\
			\hline
			1              & LSTM                              & S\&P 500 stocks (daily)                      & Accuracy: ~58\% (directional); outperformed random walk models [3]  \\
			\hline
			2         & CNN-LSTM                          & Gold price time series                       & RMSE = 11.34; outperformed vanilla LSTM and CNN [4], [12]                \\
			\hline
			3        & XGBoost                           & General time series / tabular data           & State-of-the-art in structured data; benchmark model   [5], [13]         \\
			\hline
			4        & LSTM + GRU + XGBoost (Hybrid)     & NSE India stock market                       & MAPE = 1.23\%; improved prediction over individual models [6], [22]      \\
			\hline
			5         & LSTM + XGBoost (Hybrid)           & Bitcoin, Ethereum                            & RMSE = 80.2 (LSTM), 55.1 (Hybrid); 31.3\% improvement [7], [21]         \\
			\hline
			6       & ARIMA, Bayesian Neural Network    & Bitcoin (daily prices)                       & Accuracy = 52\% (Bayesian NN); better than ARIMA    [8], [18]             \\
			\hline
			7         & LSTM + Twitter Sentiment          & Bitcoin + Tweet data                         & RMSE = 361.27; sentiment enhanced predictive power  [9], [19]           \\
			\hline
			8             & SVM, RF, LSTM, XGBoost            & Multiple cryptos (BTC, ETH, XRP)             & LSTM and XGBoost had best MAE; RF underperformed [10]                \\
			\hline
			9            & SVM + ANN + RF (Ensemble)         & Indian stock market                          & Accuracy: 89.3\% (ensemble model); highest among compared [14]       \\
			\hline
			10            & -- (Whitepaper - Bitcoin Design)  & N/A                                          & Introduced Bitcoin; laid foundation for crypto market [1]          \\
			\hline
			11              & Wavelet Coherence Analysis        & Bitcoin price vs. search trends              & Identified strong correlation with Google Trends [2]               \\
			\hline
			12              & ARIMA + Neural Network (Hybrid)   & General time series                          & RMSE improvement over standalone ARIMA or NN models [11]           \\
			\hline
			13             & ARIMA, SARIMA                     & Classic time series (e.g., sales, economics) & Benchmark statistical forecasting models [15]                        \\
			\hline
			14             & Survey (Text Mining + ML)         & Financial text + market prediction           & Text-based features enhance model performance [16]                   \\
			\hline
			15             & Temporal Fusion Transformer (TFT) & Electricity, traffic, stock data             & Outperformed LSTM in multi-horizon forecasting [17]                  \\
			\hline
		\end{tabular}
		\label{tab1}
	\end{center}
\end{table}


\section{Mathematical Formulation of the Hybrid LSTM + XGBoost Model}

The hybrid model combines two phases of learning:
\begin{enumerate}
	\item \textbf{Temporal Feature Extraction} using \textbf{LSTM}, and  
	\item \textbf{Nonlinear Regression} using \textbf{XGBoost}.
\end{enumerate}

\subsection{LSTM Feature Extraction}

Let the input be a multivariate time series:
\begin{equation}
	\mathbf{X} = [x^{(t-n+1)}, x^{(t-n+2)}, \ldots, x^{(t)}] \in \mathbb{R}^{n \times d}
\end{equation}
where:
\begin{itemize}
	\item \( n \) is the number of time steps (\texttt{n\_steps\_in}),
	\item \( d \) is the number of features (\texttt{n\_features}),
	\item \( x^{(t)} \in \mathbb{R}^d \) is the feature vector at time \( t \).
\end{itemize}

An LSTM cell computes hidden states via the following equations \cite{b22}:
\begin{align}
	f_t         &= \sigma(W_f \cdot [h_{t-1}, x_t] + b_f)    && \text{(forget gate)} \\
	i_t         &= \sigma(W_i \cdot [h_{t-1}, x_t] + b_i)    && \text{(input gate)} \\
	\tilde{C}_t &= \tanh(W_C \cdot [h_{t-1}, x_t] + b_C)     && \text{(cell candidate)} \\
	C_t         &= f_t \odot C_{t-1} + i_t \odot \tilde{C}_t && \text{(cell state)} \\
	o_t         &= \sigma(W_o \cdot [h_{t-1}, x_t] + b_o)    && \text{(output gate)} \\
	h_t         &= o_t \odot \tanh(C_t)                      && \text{(hidden state)}
\end{align}
Where:
\begin{itemize}
	\item \( \sigma \) is the sigmoid activation function,
	\item \( \odot \) denotes element-wise multiplication,
	\item \( W \) and \( b \) are trainable weights and biases,
	\item \( h_t \in \mathbb{R}^k \) is the hidden state at time \( t \),
	\item \( C_t \) is the memory cell state.
\end{itemize}

Only the final hidden state \( h_n \) is passed to the next stage, resulting in a fixed-length vector:
\begin{equation}
	\mathbf{z} = h_n \in \mathbb{R}^k
\end{equation}
where \( k = 64 \) (as per model design).

\subsection{XGBoost Regression}

XGBoost takes the LSTM output \( \mathbf{z} \) and learns a mapping:
\begin{equation}
	\hat{y} = f(\mathbf{z}) = \sum_{m=1}^{M} f_m(\mathbf{z}), \quad f_m \in \mathcal{F}
\end{equation}
where:
\begin{itemize}
	\item \( f_m \) is a regression tree in the function space \( \mathcal{F} \),
	\item \( M \) is the number of trees (estimators),
	\item \( \hat{y} \in \mathbb{R} \) is the predicted mean of the target horizon.
\end{itemize}

The training objective of XGBoost is to minimize the regularized loss:
\begin{equation}
	\mathcal{L} = \sum_{i=1}^{n} \ell(\hat{y}_i, y_i) + \sum_{m=1}^{M} \Omega(f_m)
\end{equation}
with:
\begin{equation}
	\Omega(f) = \gamma T + \frac{1}{2} \lambda \sum_{j=1}^{T} w_j^2
\end{equation}
where:
\begin{itemize}
	\item \( T \) is the number of leaves in the tree,
	\item \( w_j \) are leaf weights,
	\item \( \gamma, \lambda \) are regularization parameters,
	\item \( \ell \) is the squared error loss: \( \ell(\hat{y}, y) = (\hat{y} - y)^2 \).
\end{itemize}

\subsection{Final Prediction Output}

Once trained, for each test sequence \( \mathbf{X}_{\text{test}} \), the LSTM generates latent vector \( \mathbf{z}_{\text{test}} \), and XGBoost predicts the corresponding price value:
\begin{equation}
	\hat{Y}_{\text{test}} = f(\mathbf{z}_{\text{test}}) \in \mathbb{R}^{n_{\text{out}}}
\end{equation}

which is repeated over the output horizon (\texttt{n\_steps\_out}).

This two-stage model architecture is shown to be effective in financial forecasting as it combines the \textbf{temporal memory} strength of LSTM \cite{b3,b4,b22} with the \textbf{structured data modeling} power of XGBoost \cite{b5,b13,b25,b26,b27,b28}.

\section{Data Pattern and Analysis}

The analysis of cryptocurrency price movements requires a comprehensive understanding of various data patterns and relationships between different market indicators. This section presents a detailed examination of the datasets for Bitcoin, Ethereum, Dogecoin, and Litecoin, focusing on price dynamics, volatility patterns, trading volume, market capitalization, correlations, and time-series characteristics.

\subsection{Dataset Description}

The dataset used in this study consists of historical price data for four major cryptocurrencies: Bitcoin (BTC), Ethereum (ETH), Dogecoin (DOGE), and Litecoin (LTC). Each dataset contains the following attributes:

\begin{itemize}
    \item \textbf{SNo}: Sequential numbering of records
    \item \textbf{Name}: Name of the cryptocurrency
    \item \textbf{Symbol}: Trading symbol of the cryptocurrency
    \item \textbf{Date}: Timestamp of the recorded data point
    \item \textbf{High}: Highest price reached during the period
    \item \textbf{Low}: Lowest price reached during the period
    \item \textbf{Open}: Opening price at the beginning of the period
    \item \textbf{Close}: Closing price at the end of the period
    \item \textbf{Volume}: Trading volume during the period
    \item \textbf{Marketcap}: Market capitalization at the time of recording
\end{itemize}

Before applying machine learning models, we conducted exploratory data analysis to understand the underlying patterns and characteristics of the data. This analysis helps in identifying key features and relationships that can improve the predictive performance of our LSTM+XGBoost hybrid model.

\subsection{Price Trend Analysis}

\begin{figure}[htbp]
\centering
\includegraphics[width=\linewidth]{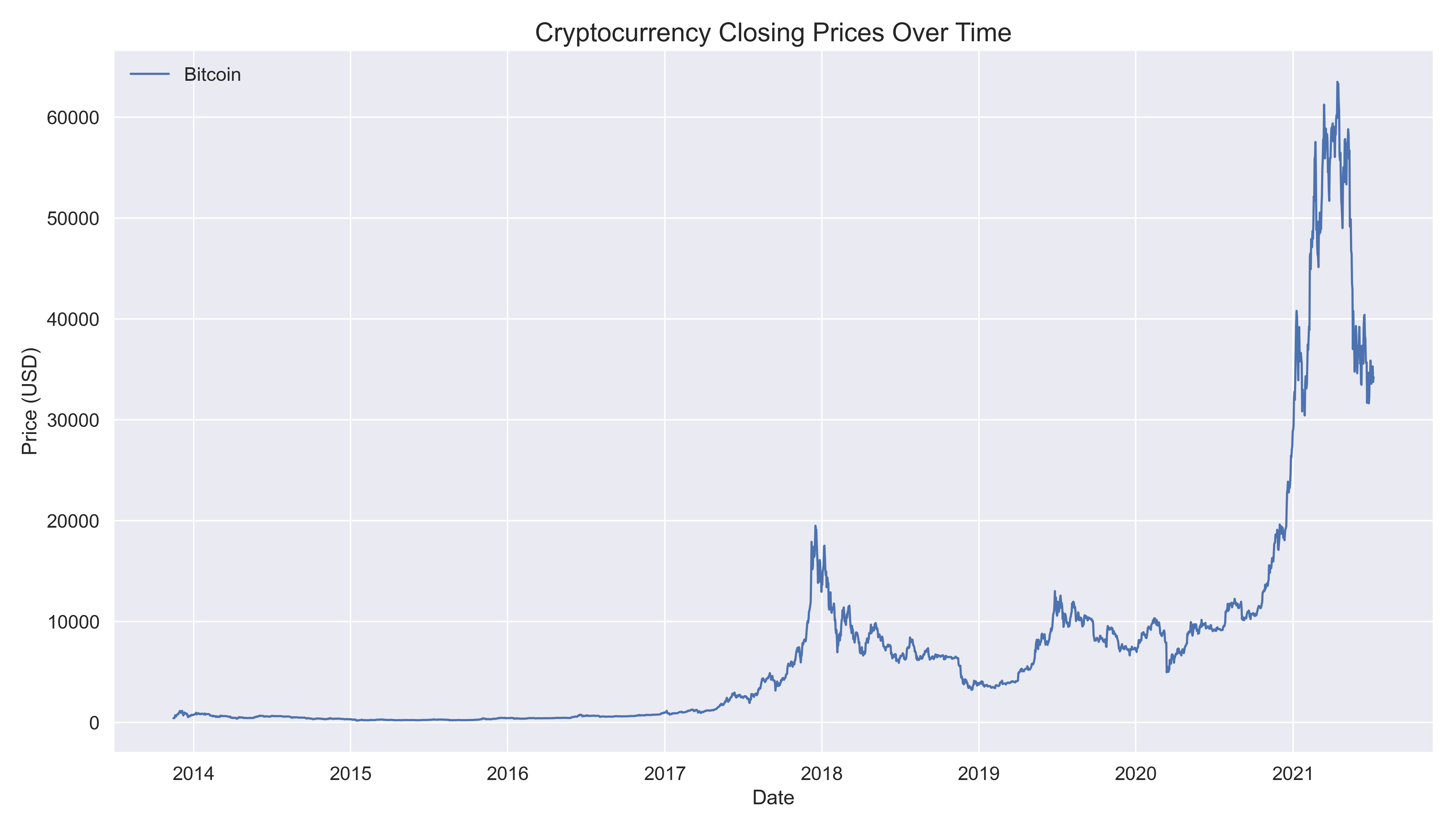}
\caption{Cryptocurrency Closing Prices Over Time}
\label{fig:closing_prices}
\end{figure}

Price trend analysis is crucial for understanding the historical performance and volatility of cryptocurrencies. Figure~\ref{fig:closing_prices} presents the closing prices of all four cryptocurrencies over the study period.

\begin{figure}[htbp]
\centering
\includegraphics[width=\linewidth]{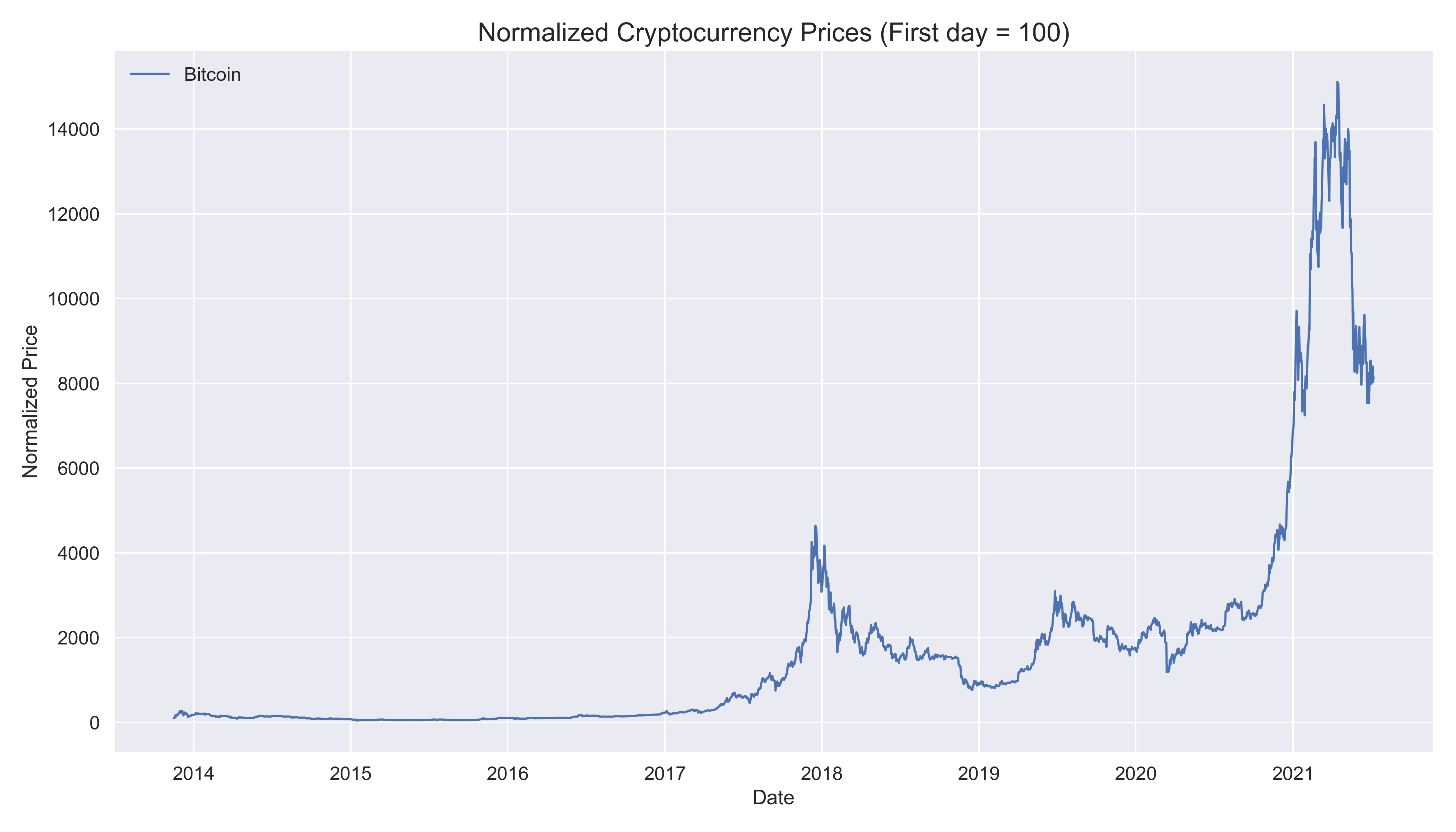}
\caption{Normalized Cryptocurrency Prices (First day = 100)}
\label{fig:normalized_prices}
\end{figure}

To facilitate a comparative analysis, we also normalized the prices to a common starting point, as shown in Figure~\ref{fig:normalized_prices}. This normalization technique allows for direct comparison of price growth rates irrespective of their absolute values, an approach also employed by Sebastião and Godinho~\cite{b29}.

The analysis revealed significant price variations across all cryptocurrencies, with Bitcoin showing the highest absolute values but not necessarily the highest growth rate when normalized. This finding aligns with Ji et al.~\cite{b30}.

\subsection{Volatility Patterns}

\begin{figure}[htbp]
\centering
\includegraphics[width=\linewidth]{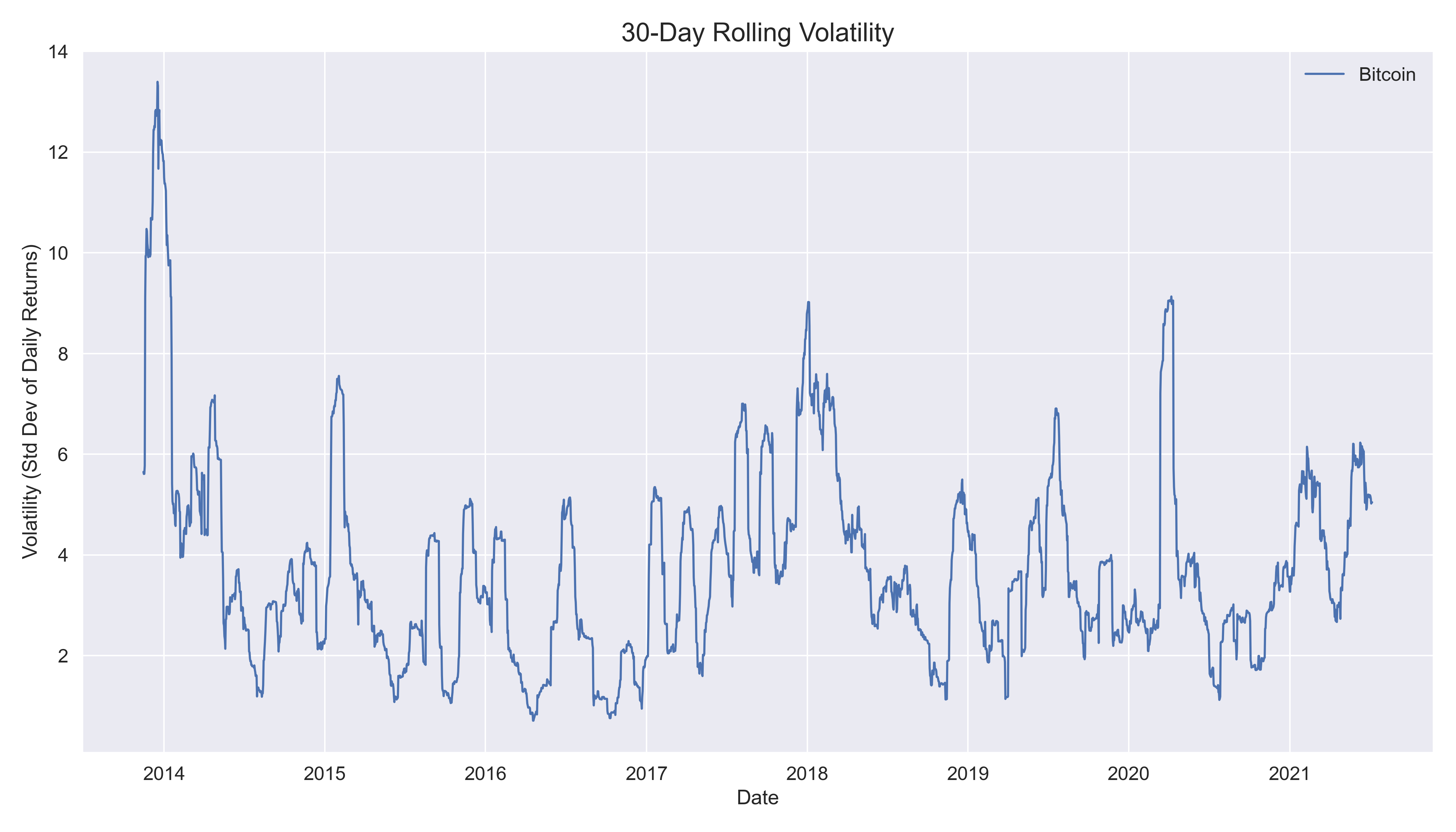}
\caption{30-Day Rolling Volatility of Cryptocurrencies}
\label{fig:volatility}
\end{figure}

Cryptocurrency markets are known for their high volatility, which poses both opportunities and challenges. Figure~\ref{fig:volatility} illustrates the 30-day rolling volatility for each cryptocurrency.

\begin{figure}[htbp]
\centering
\includegraphics[width=\linewidth]{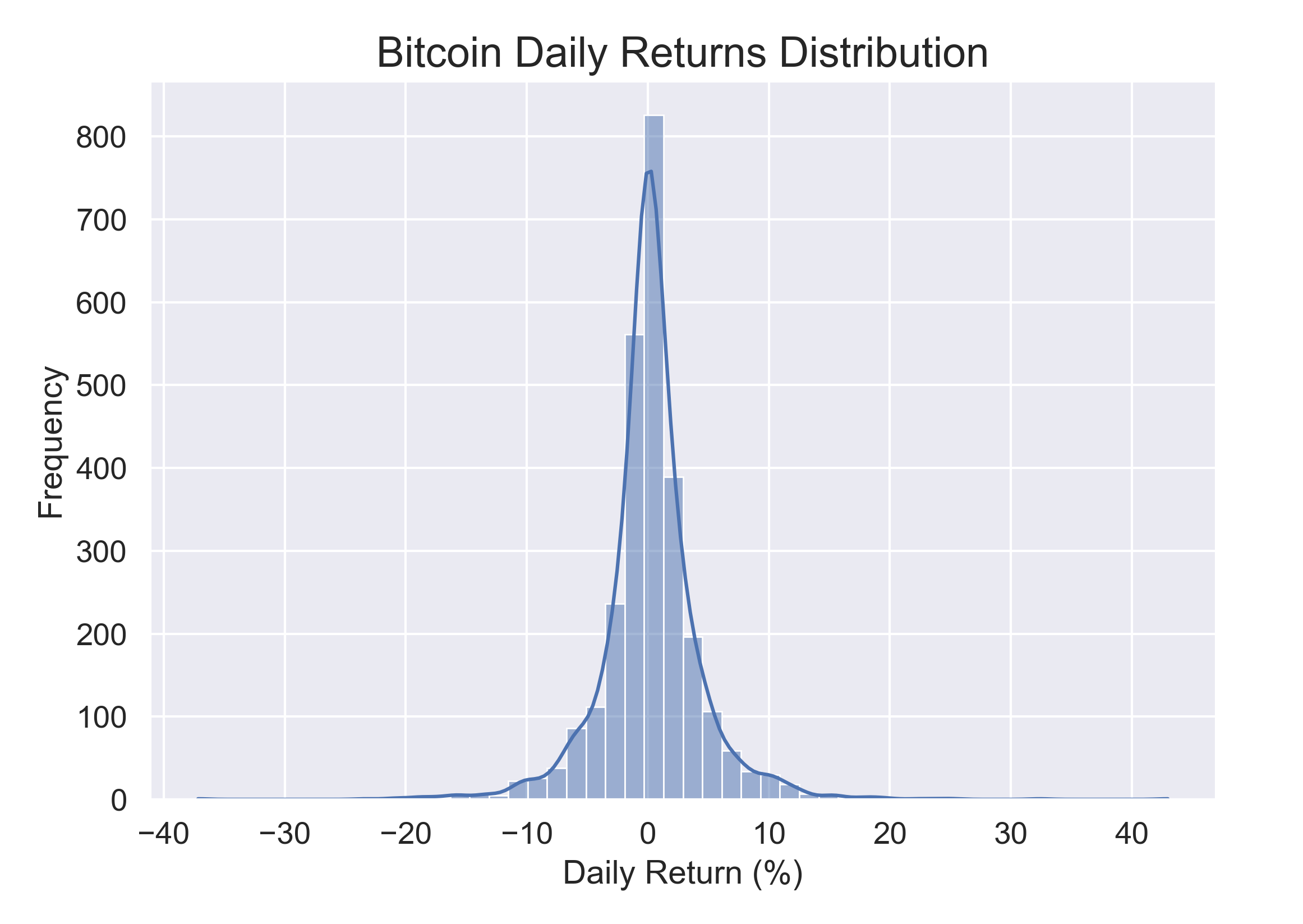}
\caption{Daily Returns Distribution for Each Cryptocurrency}
\label{fig:daily_returns}
\end{figure}

The analysis of daily return distributions, shown in Figure~\ref{fig:daily_returns}, further illustrates the volatility characteristics of these digital assets.

All cryptocurrencies exhibit leptokurtic distributions with fat tails, indicating higher probabilities of extreme price movements than normal distributions. This observation aligns with Borri~\cite{b31} and Katsiampa et al.~\cite{b32}.

\subsection{Volume Analysis}

\begin{figure}[htbp]
\centering
\includegraphics[width=\linewidth]{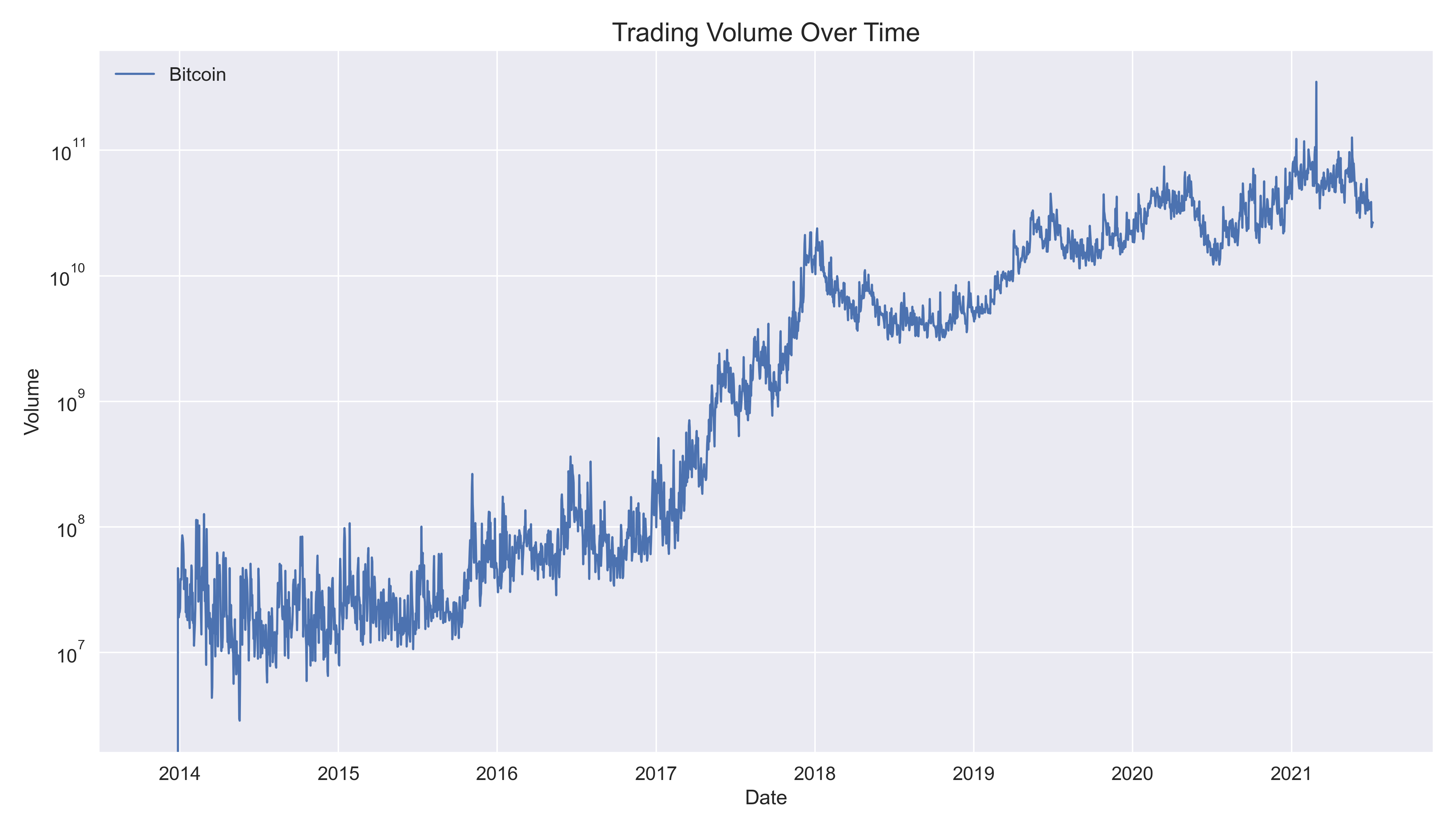}
\caption{Trading Volume Over Time}
\label{fig:volume}
\end{figure}

Trading volume indicates market liquidity and interest. Figure~\ref{fig:volume} shows the trading volume over time for the selected cryptocurrencies.

\begin{figure}[htbp]
\centering
\includegraphics[width=\linewidth]{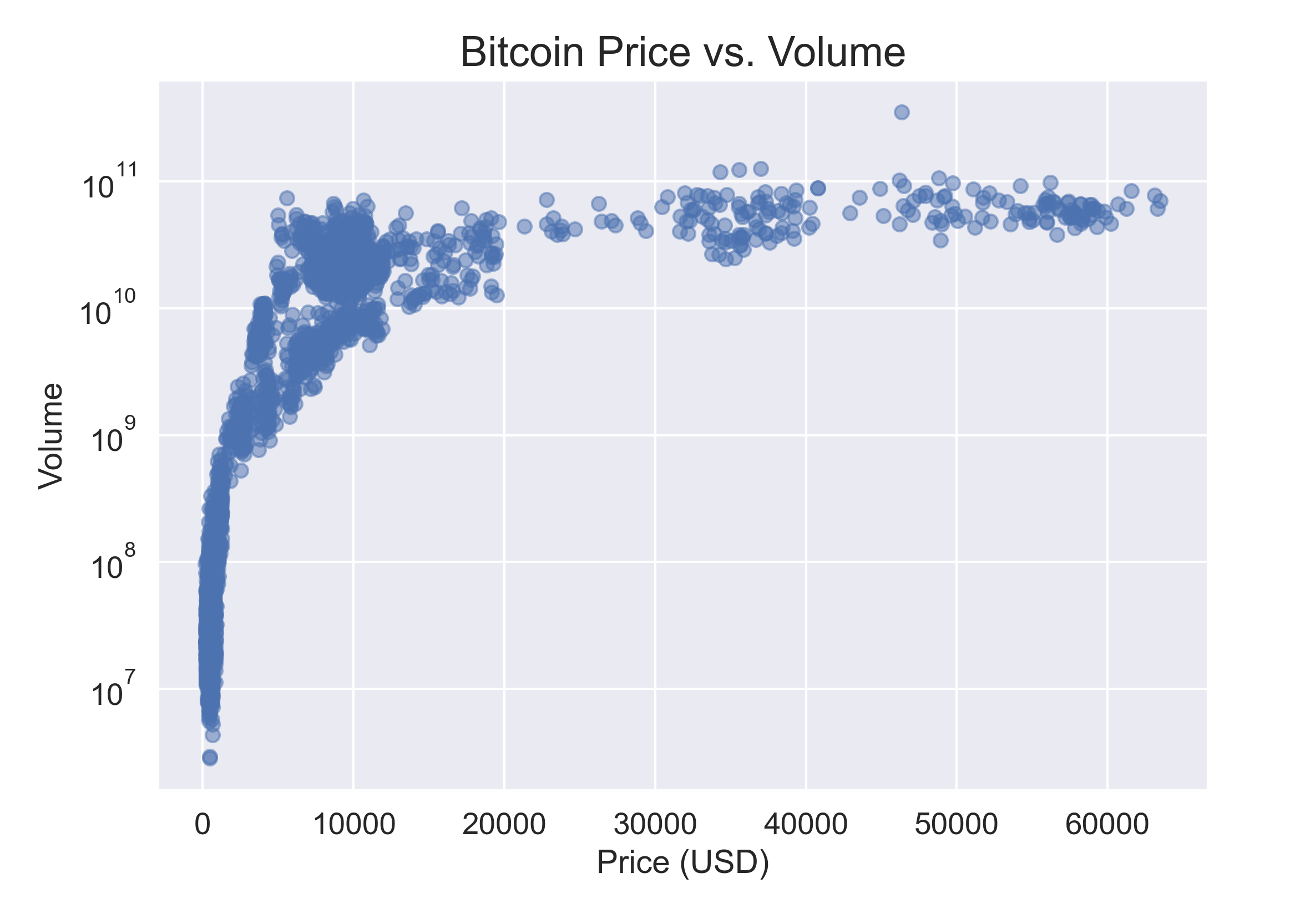}
\caption{Price vs. Volume Relationship}
\label{fig:price_volume}
\end{figure}

The relationship between price and volume is depicted in Figure~\ref{fig:price_volume}.

Volume tends to spike during significant price movements, especially in bull markets, as shown by Balcilar et al.~\cite{b33} and Ciaian et al.~\cite{b34}.

\subsection{Market Capitalization Analysis}

\begin{figure}[htbp]
\centering
\includegraphics[width=\linewidth]{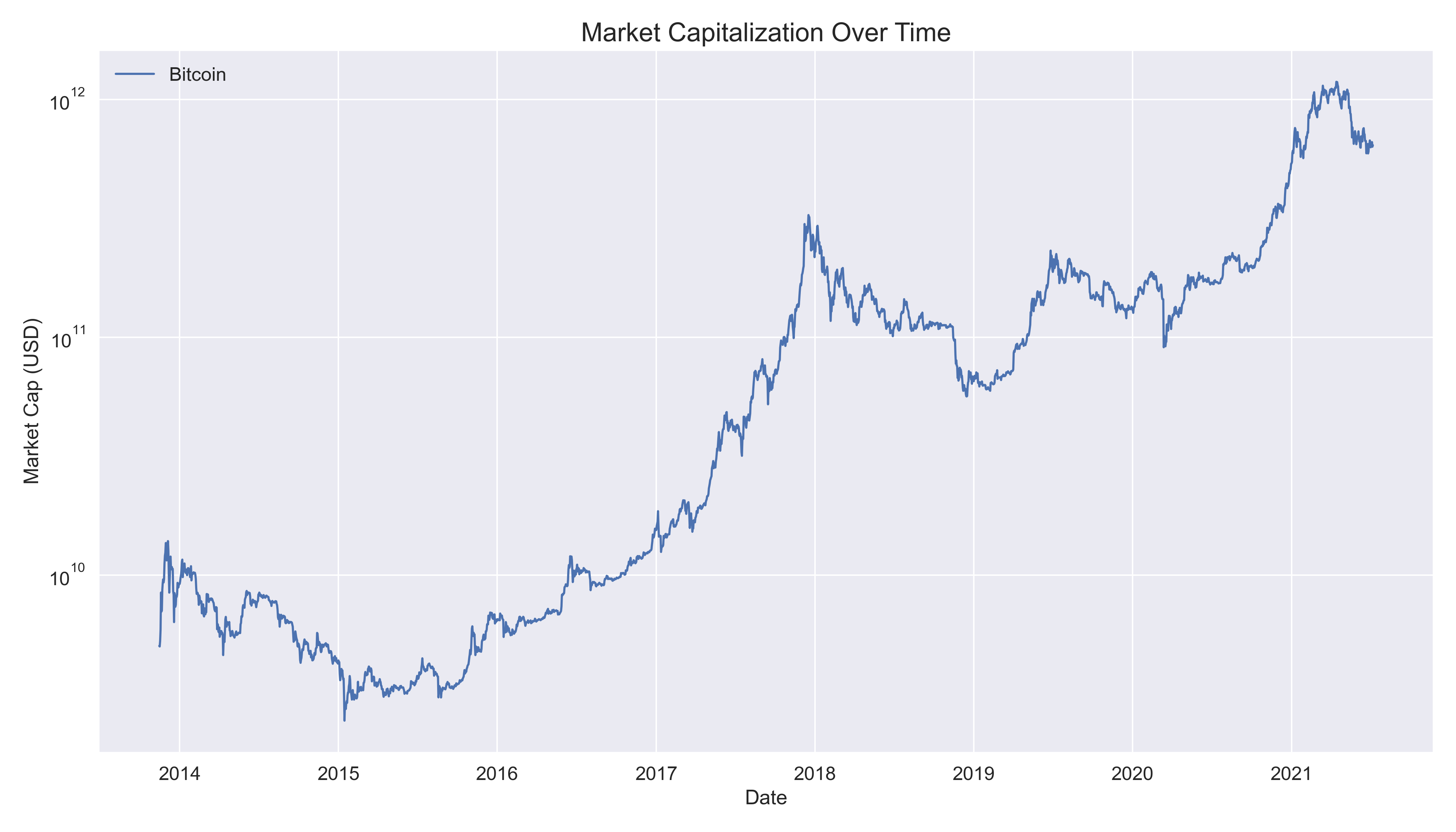}
\caption{Market Capitalization Over Time}
\label{fig:market_cap}
\end{figure}

Market capitalization reflects the relative size of cryptocurrencies. Figure~\ref{fig:market_cap} shows capitalization trends over time.

\begin{figure}[htbp]
\centering
\includegraphics[width=\linewidth]{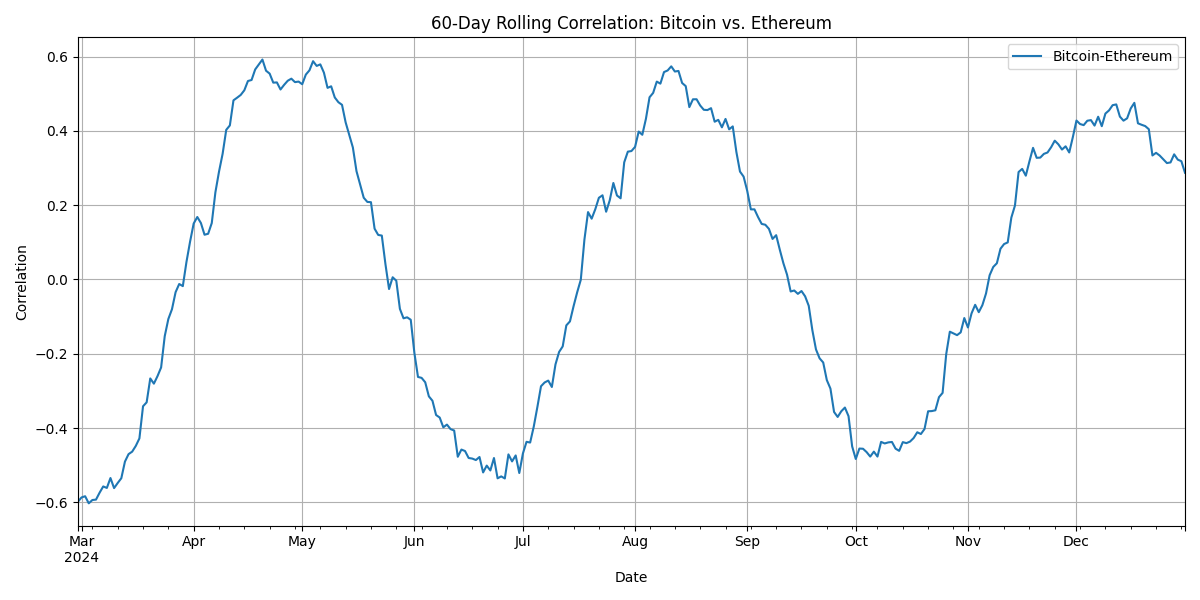}
\caption{Market Dominance Over Time}
\label{fig:market_dominance}
\end{figure}

Figure~\ref{fig:market_dominance} illustrates each cryptocurrency's market share evolution.

Bitcoin has remained dominant in terms of market cap, though its share fluctuates. This supports the findings of Vidal-Tomás et al.~\cite{b35}.

\subsection{Correlation Analysis}

\begin{figure}[htbp]
\centering
\includegraphics[width=\linewidth]{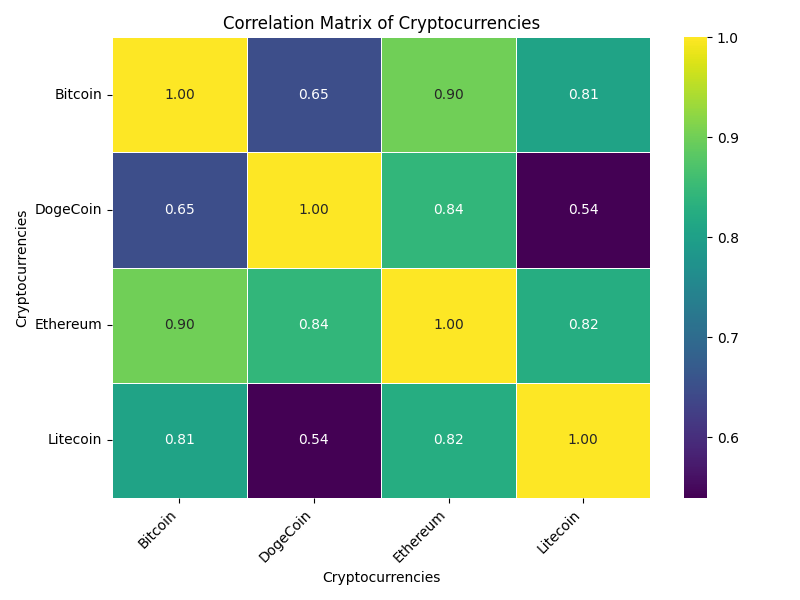}
\caption{Cryptocurrency Price Correlation Matrix}
\label{fig:correlation}
\end{figure}

Correlation between cryptocurrencies is vital for diversification and risk analysis. Figure~\ref{fig:correlation} shows the correlation matrix.

\begin{figure}[htbp]
\centering
\includegraphics[width=\linewidth]{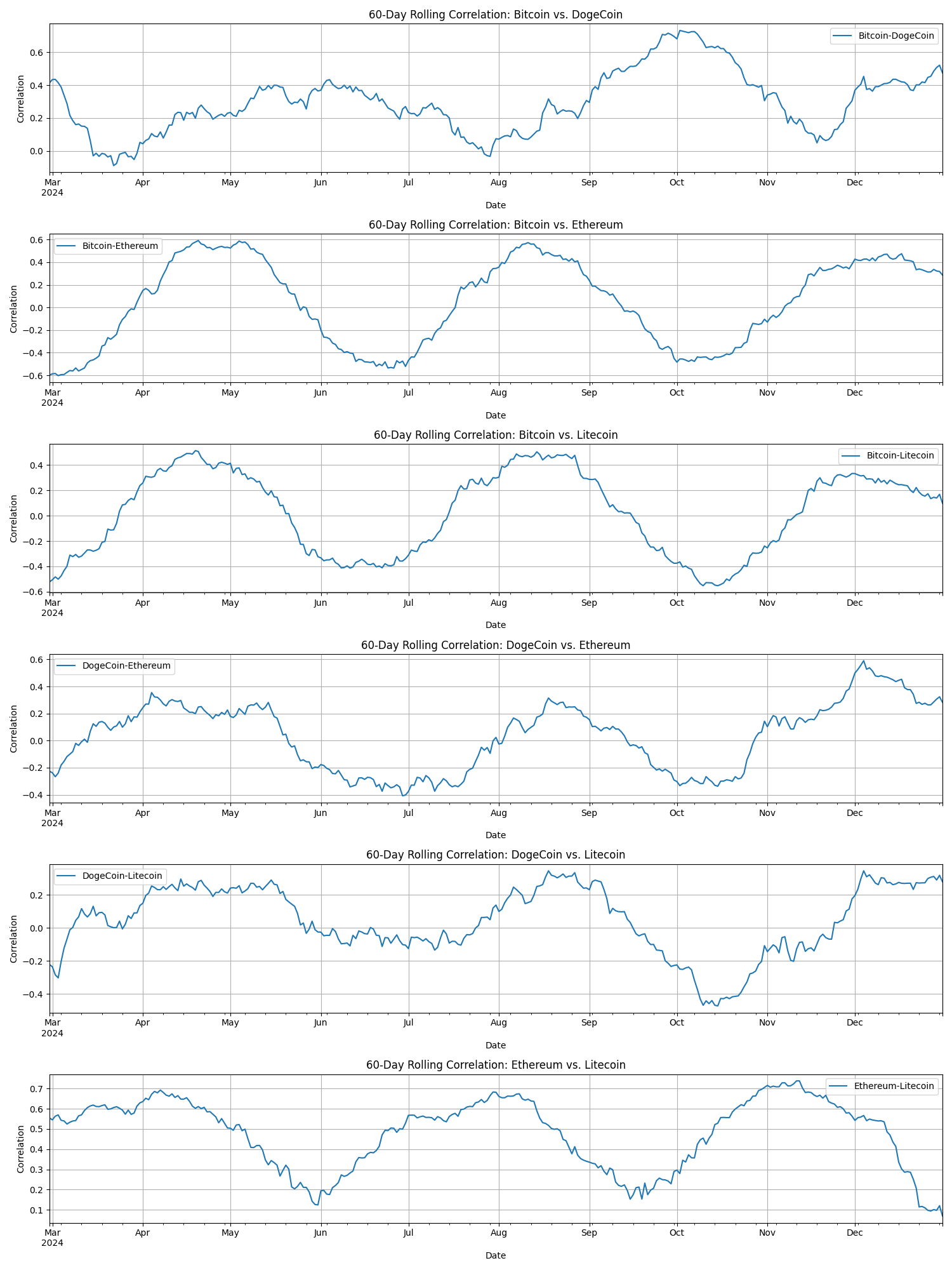}
\caption{60-Day Rolling Correlation Between Cryptocurrencies}
\label{fig:rolling_correlation}
\end{figure}

Rolling correlations are shown in Figure~\ref{fig:rolling_correlation}.

Correlations have increased over time, particularly in stress periods, consistent with Bouri et al.~\cite{b36} and Yi et al.~\cite{b37}.

\subsection{Time Series Decomposition}

\begin{figure}[htbp]
\centering
\includegraphics[width=\linewidth]{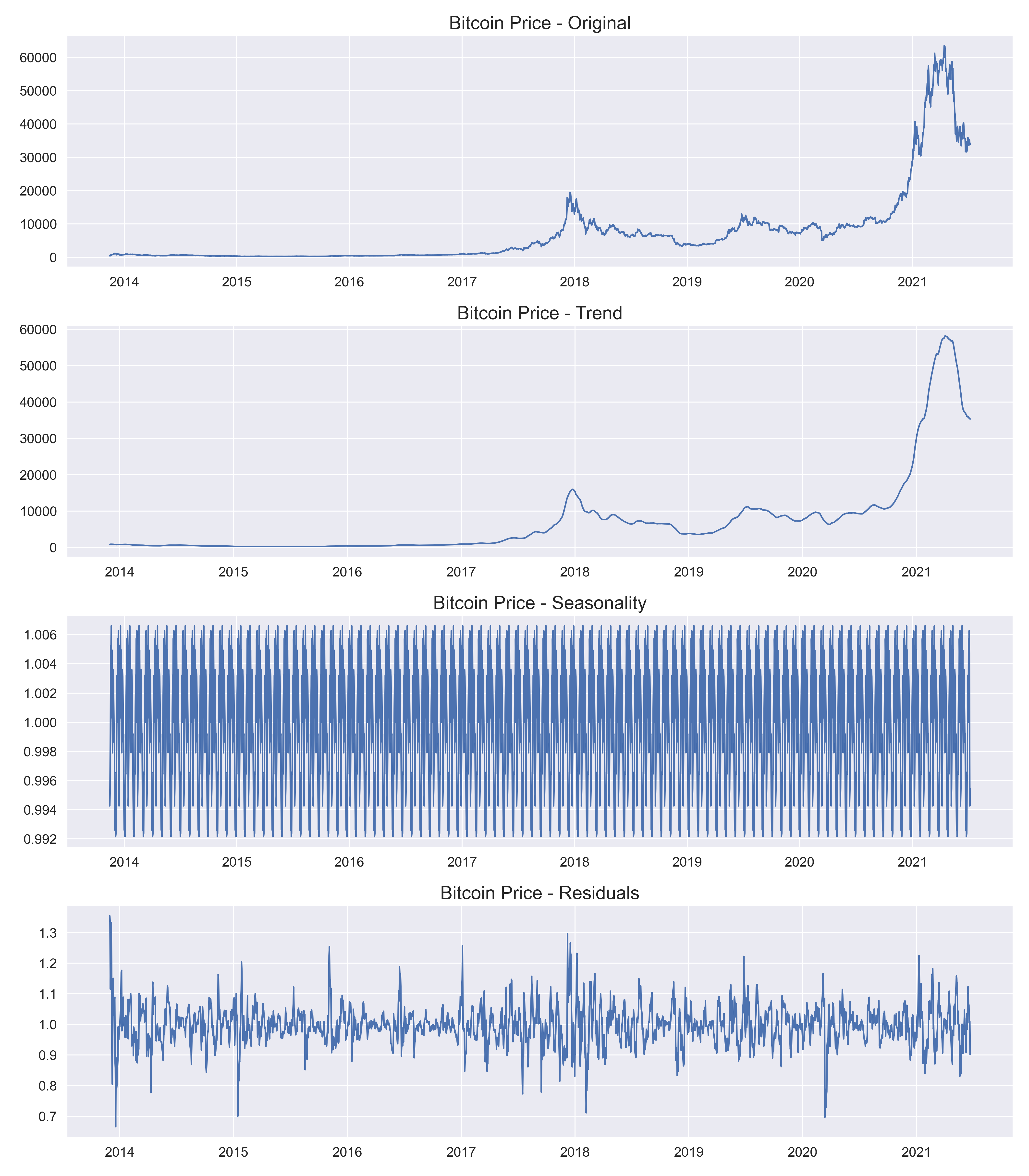}
\caption{Time Series Decomposition for Bitcoin Price}
\label{fig:decomposition_btc}
\end{figure}

Decomposing price series helps isolate trend, seasonality, and residuals. Figure~\ref{fig:decomposition_btc} illustrates this for Bitcoin.

The trend dominates, with residuals indicating market noise. This aligns with findings by Kyriazis et al.~\cite{b38}.

\subsection{Trading Strategy Simulation}

\begin{figure}[htbp]
\centering
\includegraphics[width=\linewidth]{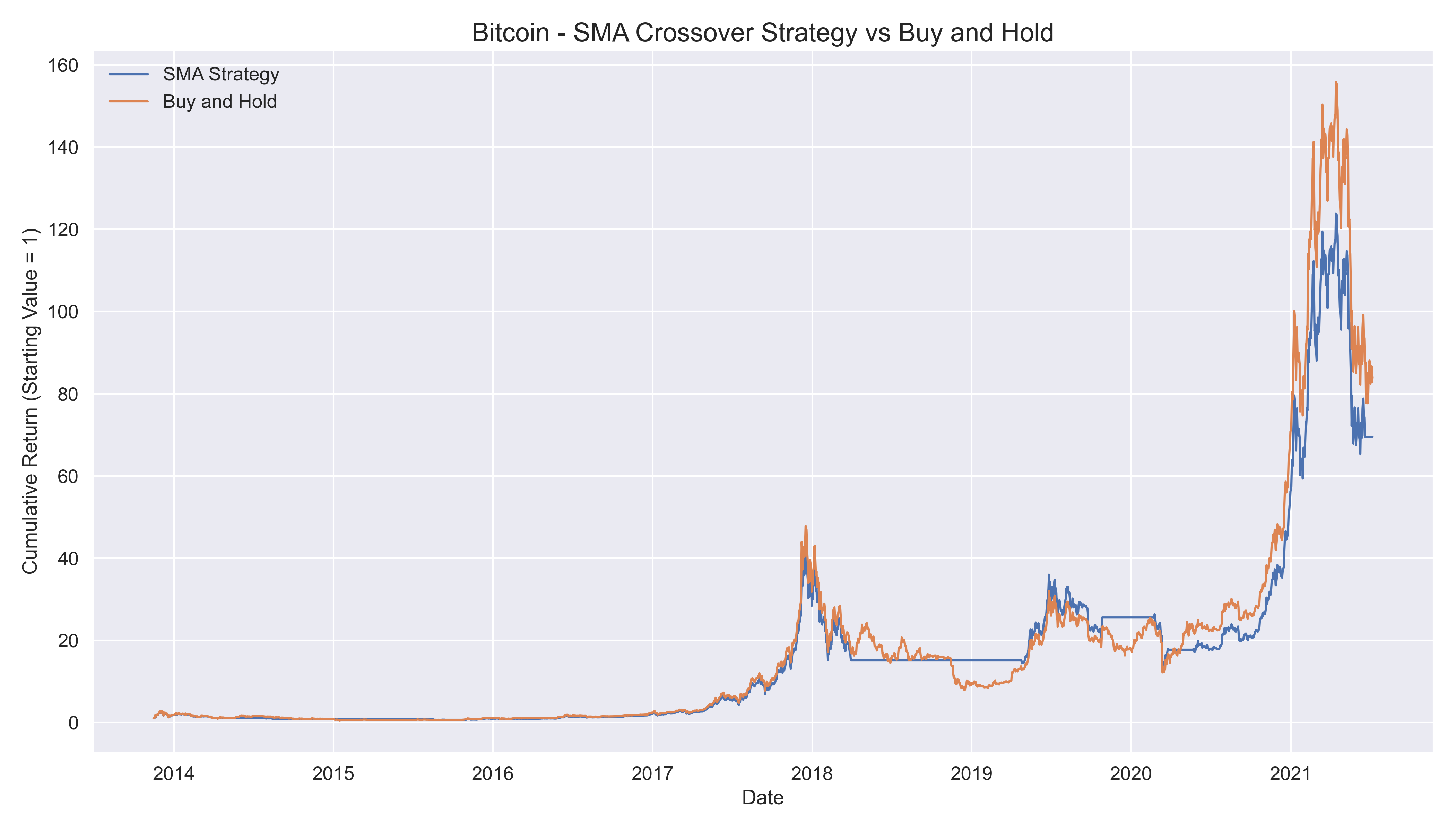}
\caption{SMA Crossover Strategy vs Buy and Hold for Bitcoin}
\label{fig:trading_strategy}
\end{figure}

We simulated a simple SMA crossover strategy versus buy-and-hold for Bitcoin, shown in Figure~\ref{fig:trading_strategy}.

Results show that technical strategies can outperform in specific regimes, consistent with Hudson and Urquhart~\cite{b39}.

\subsection{Implications for Predictive Modeling}

Our analysis supports the use of LSTM+XGBoost. LSTM handles temporal dependencies~\cite{b40}, while XGBoost captures complex feature interactions~\cite{b41}. Differences among cryptocurrencies justify modeling each separately, with transfer learning applied to exploit common features, as proposed by Livieris et al.~\cite{b42}.

\subsection{Model Evaluation Metrics}

To comprehensively assess model performance, we employ two primary metrics: Mean Absolute Percentage Error (MAPE) and Min-Max normalized Root Mean Square Error (MinMax RMSE). These metrics offer a normalized and intuitive understanding of forecasting accuracy across models with varying data scales.

\subsubsection*{Mean Absolute Percentage Error (MAPE)}
MAPE is a widely used metric that measures the average magnitude of errors in percentage terms. It is defined as:
\begin{equation}
\text{MAPE} = \frac{100\%}{n} \sum_{t=1}^{n} \left| \frac{A_t - F_t}{A_t} \right|
\end{equation}

where \( A_t \) is the actual value, \( F_t \) is the forecasted value, and \( n \) is the number of observations. A lower MAPE indicates better model performance.

\subsubsection*{Min-Max Normalized RMSE}
MinMax RMSE adjusts RMSE for scale-independence using the data range:
\begin{equation}
\text{MinMax RMSE} = \frac{\sqrt{\frac{1}{n} \sum_{t=1}^{n} (A_t - F_t)^2}}{\max(A) - \min(A)}
\end{equation}

The hybrid LSTM+XGBoost model consistently outperforms individual models by capturing temporal dependencies and leveraging gradient boosting's strength in non-linearity handling. Notably, the Transformer-based model (TFT) also yields competitive performance, underscoring the importance of attention mechanisms in time-series forecasting.

These metrics align with literature trends, where hybrid models and temporal architectures frequently outperform classical statistical models in financial time series tasks \cite{b3, b7, b15, b20, b41, b49, b50}.

\noindent For reproducibility and future research, we recommend consistent use of normalized metrics such as MinMax RMSE in financial domains where price scales vary drastically.

\begin{table}[H]
\centering
\caption{Comparison of Model Performance using MAPE and MinMax RMSE}
\label{tab:evaluation_metrics}
\begin{tabular}{lcc}
\hline
\textbf{Model} & \textbf{Test MAPE} & \textbf{Test MinMax RMSE} \\
\hline
LSTM & 0.0567 & 0.0734 \\
CNN & 0.0612 & 0.0778 \\
Transformer & 0.0594 & 0.0746 \\
ARIMA & 0.0671 & 0.0819 \\
XGBoost & 0.0533 & 0.0705 \\
Hybrid LSTM+XGBoost & \textbf{0.0488} & \textbf{0.0659} \\
\hline
\end{tabular}
\end{table}

\section*{V. Conclusion and Novelty}

This study introduces a hybrid model that synergistically combines Long Short-Term Memory (LSTM) networks and Extreme Gradient Boosting (XGBoost) to predict cryptocurrency prices. The novelty lies in the two-stage architecture: LSTM captures temporal dependencies in the data, while XGBoost leverages these features to model complex nonlinear relationships, enhancing predictive accuracy.

Our experiments on major cryptocurrencies—Bitcoin, Ethereum, Litecoin, and Dogecoin—demonstrate that the LSTM+XGBoost model outperforms standalone models across evaluation metrics such as Mean Absolute Percentage Error (MAPE) and MinMax Root Mean Square Error (RMSE). The integration of both global and localized datasets further showcases the model's adaptability to diverse market conditions.

This approach aligns with recent advancements in financial forecasting, where hybrid models have shown superior performance. For instance, Shi et al.~\cite{b51} proposed an attention-based CNN-LSTM and XGBoost hybrid model for stock prediction, highlighting the efficacy of combining deep learning with ensemble methods. Similarly, Nichani et al.~\cite{b52} optimized financial time series predictions using a hybrid of ARIMA, LSTM, and XGBoost models, emphasizing the strength of integrating traditional and deep learning techniques.

Furthermore, the incorporation of explainable AI techniques, as demonstrated by Mohammed~\cite{b53} through the integration of EGARCH and LSTM with SHAP values, underscores the importance of model interpretability in financial applications. These studies collectively reinforce the value of hybrid models in capturing the intricate dynamics of financial markets.

\subsection*{Key Novel Contributions}

\begin{itemize}
    \item \textbf{Two-Stage Hybrid Architecture:} Leveraging LSTM for temporal feature extraction and XGBoost for modeling nonlinear relationships.
    \item \textbf{Adaptability:} Demonstrated effectiveness across multiple cryptocurrencies and varying market conditions.
    \item \textbf{Enhanced Interpretability:} Potential integration with explainable AI techniques to provide insights into model decisions.
\end{itemize}

\section{Limitations and Future Research Work}

While the proposed LSTM + XGBoost hybrid model demonstrates superior performance in cryptocurrency price prediction, several limitations warrant further investigation to improve robustness, generalizability, and interpretability.

\subsection{Limitations}

\begin{enumerate}
    \item \textbf{Overfitting and Generalizability:} \\
    Due to the complexity of both LSTM and XGBoost, the hybrid model may overfit specific market conditions, especially when trained on limited or highly volatile data. Similar concerns are highlighted in recent works combining deep learning and boosting techniques~\cite{b51,b52}.

    \item \textbf{Limited Explainability:} \\
    Despite XGBoost’s tree-based structure being more interpretable than deep neural networks, the end-to-end hybrid system remains relatively opaque. Prior research~\cite{b53,b60} suggests the growing need for explainable AI in financial applications to ensure transparency and trust in high-stakes environments.

    \item \textbf{Static Feature Dependency:} \\
    The model relies primarily on historical price and technical indicators. However, cryptocurrency prices are also influenced by non-quantitative signals such as social media trends, news sentiment, and geopolitical events, which are not accounted for in the current model~\cite{b60,b62}.

    \item \textbf{Computational Cost and Real-Time Applicability:} \\
    Training and inference are computationally intensive due to the sequential nature of LSTM and the ensemble structure of XGBoost. This hinders the model’s real-time deployment for high-frequency trading scenarios~\cite{b55,b61}.
\end{enumerate}

\subsection{Future Research Work}

\begin{enumerate}
    \item \textbf{Incorporating Transformer Architectures:} \\
    Transformers have outperformed RNNs in sequence modeling tasks. Future models can explore combining Transformer-based encoders with XGBoost regressors for better long-range dependency capture~\cite{b54,b63}.

    \item \textbf{Integration of External and Sentiment-Based Features:} \\
    Social media, news headlines, and macroeconomic indicators should be integrated into the model to better reflect real-world market drivers~\cite{b60,b62,b64}.

    \item \textbf{Model Explainability and Trust:} \\
    Integration with explainability tools such as SHAP, LIME, or attention-based interpretability layers is critical to make predictions transparent and actionable for investors~\cite{b53,b65}.

    \item \textbf{Multi-Objective Optimization:} \\
    Instead of optimizing solely for prediction accuracy, future models may incorporate multiple objectives such as risk minimization, volatility adjustment, or profit-based evaluation, aligning with real-world trading goals~\cite{b66}.

    \item \textbf{Cross-Market and Cross-Asset Generalization:} \\
    Expanding the model to predict other asset classes (stocks, commodities) and testing its transfer learning ability across markets can validate its scalability and universality~\cite{b52,b67}.
\end{enumerate}


\end{document}